\documentclass{article}
\usepackage{spconf,amsmath,graphicx}
\newcommand{\etal}{\textit{et al.}}
\DeclareMathOperator{\st}{s.t.}
\usepackage[linesnumbered,lined,boxed,commentsnumbered,ruled,vlined,]{algorithm2e}

\title{One-Shot Neural Band Selection for Spectral Recovery}
\name{Hai-Miao Hu$^{1,4}$,Zhenbo Xu$^{1}$,Wenshuai Xu$^{2\star}$\thanks{
This work was partially supported by the "Pioneer" and "Leading Goose" R\&D Program of Zhejiang (Grant No. 2022C01082), the National Natural Science Foundation of China (No.62206012, No.62122011, U21A20514), and the China Postdoctoral Science Foundation (Grand No. 2021M700346), and the Natural Science Foundation of Zhejiang province (No. Q23F020065), and the Fundamental Research Funds for the Central Universities.
$^{\star}$ Corresponding author. Email: xuzhenbo@mail.ustc.edu.cn, xu@buaa.edu.cn
},You Song$^{2}$,YiTao Zhang$^{2}$,Liu Liu$^{3}$,Zhilin Han$^{3}$,Ajin Meng$^{3}$}
\address{\textsuperscript{1}Hangzhou Innovation Institute, Beihang University, Hangzhou, China  \\
\textsuperscript{2}School of Software, Beihang University \\
\textsuperscript{3}ShiFang Technology Inc., Hangzhou, China \\
\textsuperscript{4}State Key Laboratory of Virtual Reality Technology and Systems, Beihang University}

\begin{document}
%
\maketitle
\begin{abstract}
Band selection has a great impact on the spectral recovery quality. To solve this ill-posed inverse problem, most band selection methods adopt hand-crafted priors or exploit clustering or sparse regularization constraints to find most prominent bands. These methods are either very slow due to the computational cost of repeatedly training with respect to different selection frequencies or different band combinations. Many traditional methods rely on the scene prior and thus are not applicable to other scenarios. In this paper, we present a novel one-shot Neural Band Selection (NBS) framework for spectral recovery. Unlike conventional searching approaches with a discrete search space and a non-differentiable search strategy, our NBS is based on the continuous relaxation of the band selection process, thus allowing efficient band search using gradient descent. To enable the compatibility for selecting any number of bands in one-shot, we further exploit the band-wise correlation matrices to progressively suppress similar adjacent bands. Extensive evaluations on the NTIRE 2022 Spectral Reconstruction Challenge demonstrate that our NBS achieves consistent performance gains over competitive baselines when examined with four different spectral recovery methods. Our code will be publicly available.

\end{abstract}
\section{Introduction}
\label{sec:intro}

In coded aperture snapshot spectral imaging (CASSI) system \cite{quan2022high}, spectral recovery is proposed to recover a hyperspectral image from its 2D snapshot measurements that can be captured by more cheaper and faster snapshot multispectral imaging systems. 
To enable effective spectral recovery, band selection (BS) algorithms \cite{giannopoulos20224d,zhang2022graph,xiong2022multitask,li2022material,lee2022self,han2022spatial,hu2022hdnet} are designed to select most informative bands to alleviate the difficulty of the ill-posed problem. Effective and efficient BS is essential for wide applications of hyperspectral image sensing because it enables the fast discovery of a limited number of prominent bands to design intelligent multi-spectral imaging systems that are much cheaper and can operate in real-time.

Current band selection approaches \cite{sun2019hyperspectral} can be divided into supervised and unsupervised according to scenarios and the existence of labels. Supervised methods \cite{zhang2022graph} search optimal bands by training models with task-specific objectives on the labelled data set. Feng \etal \cite{yang2010efficient} proposed a supervised band selection method by exploiting the known class signatures. In \cite{feng2017hyperspectral}, to construct effective band selections, authors presented a new pointwise-ranking-based band selection paradigm by employing a non-homogeneous hidden Markov chain model and known labels. Recently, deep learning based methods show great potential for band selection. Feng \etal \cite{feng2021dual} designed a novel dual-graph neural network with attention and sparse constraint for band selection. To eliminate iterative training, recent methods further incorporate deep reinforcement learning \cite{mou2021deep} and graph learning based autoencoder \cite{zhang2022graph} for band selection. Unsupervised methods \cite{yang2010efficient,nandi2023tattmsrecnet} adopt graphs or clustering methods and exploit both spatial features and spectral features \cite{zhang2022robust} to achieve robust spectral band selection. Though current band selection methods achieve promising results, iterative training on different band selection choices is usually needed to find out an outstanding result. Besides, when the number of bands to select changes, the search procedures (especially for clustering-based methods \cite{wang2018optimal}) need to restart from scratch.

In this paper, we proposed an effective one-shot band selection framework coined Neural Band Selection (NBS) for spectral recovery. Unlike conventional searching approaches with a discrete search space and a non-differentiable search strategy, inspired by recent advances in neural architecture search \cite{liu2018darts}, our NBS is based on the continuous relaxation of the band selection process, thus allowing efficient band search using gradient descent.
Though our flexible framework can be generalized to other hyperspectral analysis tasks by applying minor changes to the loss function, to keep this paper focused, we only examine its effectiveness and flexibility on spectral recovery. 
The performance of NBS is extensively examined on the NTIRE 2022 spectral reconstruction challenge with four different spectral recovery methods (MST++ \cite{cai2022mst++}, MST-L \cite{cai2022mask}, MIRNet \cite{zamir2020learning}, HINet \cite{hu2022hdnet}).
Results demonstrate that our NBS searches better band combinations than strong baselines as well as manual selections, and can select different number of bands in one-shot.


\begin{figure*}[]
\centering
\includegraphics[width=0.58\linewidth]{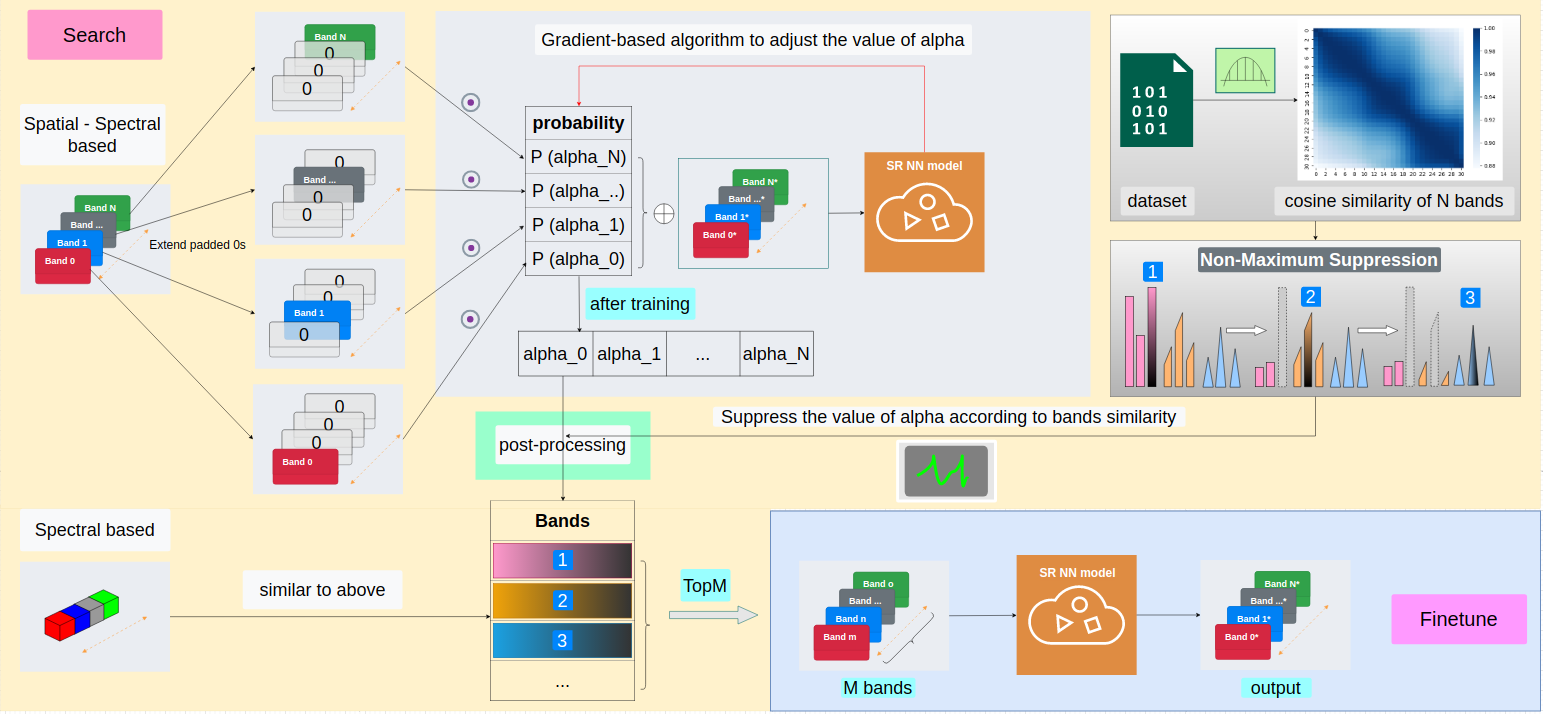}
\caption{Overview of our Neural Band Search. NBS makes the discrete band search continuous by relaxing the selection operation to a softmax of all selection operations. Then, various number of prominent bands can be inferred in one-shot by a simple post-processing algorithm.
}
\label{method_fig}
\end{figure*}

\section{Method}
In this section, we present the formulations of band selection and spectral recovery. Then, the framework of our method including the continuous relaxation of band selection is introduced. Lastly, we discuss two variants of our NBS to explore the impact of spatial features as well as the global search strategy on the design of our NBS.




\subsection{Problem statement}
Let $N$ represent the number of bands in hyperspectral images and $I=\{B^{(u,v)}_i, i\in (1, N), u \in (1, W), v \in (1,H)\} \in R^{H\times W\times B}$ denote each hyperspectral image $I$ consists of $N$ images. The height $H$ and the width $W$ of each image are two spatial dimensions and the band index $i$ is the spectral dimension. $B^{(u,v)}_i$ represent the pixel at the coordinate $(u, v)$ on the band image $B_i$. The spectral-wise input sequence at the coordinate $(u, v)$ can be represented as $S(u,v) = \{B^{(u,v)}_i, i\in (1, N)\}$.
Let $E$ denote the spectral recovery model. $M$ is the number of selected bands for spectral recovery and $M$ satisfies $M < N$.
The spectral recovery aims to recover the full hyperspectral image $I$ from limited $M$ spectral bands and can be formulated as:
\begin{equation}
	\hat{I} = G(\sum_{i=1}^{M} B^{(u,v)}_{b_i}), b_i \in (1, N)
\end{equation}
where $\{b_i, i\in (1, M)\}$ denote the selected spectral bands and G present the spectral recovery model. The number of input spectral bands $M$ has a great impact on the upper bound of the spectral recovery quality. As shown in Table \ref{ablation_study}, when we recover hyperspectral image from $M=3$ bands rather than $M=2$ bands, the PSNR increases significantly by 18.8\%.

The training objective of spectral recovery $L_r$ is commonly formulated as the distance between $I$ and $\hat{I}$. Following the NTIRE 2022 Spectral Challenge \cite{arad2022ntire}, the mean relative absolute error (MRAE) is adopted by default.

\subsection{Neural Band Search}
\label{nbs_subsection}
Though the number of input spectral bands is important for good spectral recovery quality, another essential factor is the chosen of informative spectral bands. Here we introduce a simple but effective band search method. Though we focus on spectral recovery in this paper, we believe our method is also applicable to band search for other hyperspectral analysis tasks with minor changes on the learning objective.

The band selection is inherently a discrete process.
As shown in Fig. \ref{method_fig}, our NBS makes the discrete search strategy continuous by first applying the spectral padding operation and then relaxing the band selection operations to a softmax of all selection operations. Different from previous band search method \cite{wang2018optimal}, we directly input all spectral bands and learn the weight of each spectral band by designing the relaxed selection operator $\alpha$.
Let $\{B_i\}^{N}_{i=1}$ present the input spectral image. First, NBS apply the spectral padding operation $D$ to expand each spectral band $B_{i} \in R^{H \times W}$ to $P_{i}=D(B_{i}) \in R^{H \times W \times N}$ with the channels of other $N-1$ bands filled with zero.
After that, each $\{P_{i}, i \in (1, N)\}$ has the same shape $H \times W \times N$ as the original hyperspectral image. The mixed operation $o^i$ applied to the padded spectral image $\{P_i\}^{N}_{i=1}$ as follows:

\begin{equation}
	o^i(B_i) = \sum_{o \in O}\frac{\exp (\alpha_{o}^{i})}{\sum_{o^{'} \in O} \exp (\alpha_{o^{'}}^{i})} o(B_i)
\label{eq_darts}
\end{equation}
where $O$ denotes a set of band selection operations, while $\alpha_o^i$ represents the weight of each operation $o$ on selecting the spectral band $B_i$. Therefore, the band selection has evolved into an optimization process for a set of continuous variables $\alpha = \{\alpha^i\}$, which represent the priority of each spectral band. In addition, we apply a relatively strong $L2$ regularization with a weight value of $0.01$ on $\alpha$ in training to control its sparsity. 

Let $L_r^T$ denote the training loss and $L_r^V$ represent the validation loss. Given the weight $w$ of the spectral recovery model, the search process is a bilevel optimization problem \cite{liu2018darts} as follows:
\begin{equation}
    \min_{\alpha} L_r^V (w^{\star} (\alpha), \alpha),
    \st \quad w^{\star} (\alpha) = \text{argmin}_w L_r^T(w, \alpha)
\label{darts_optim}
\end{equation}
Our NBS finds $\alpha$ that minimizes the validation loss $L_r^V (w^{\star}, \alpha)$, where $w^{\star}$ is computed by minimizing the training loss $w^{\star} = \text{argmin}_w L_r^T(w, \alpha)$.

Once the search is complete, the most prominent $M$ spectral bands are selected based on the jointly considering $\alpha$ and the band-wise correlation matrices $C \in R^{N \times N}$. $C$ equals to the cosine similarity between each spectral band by averaging all pixels in the training set. The larger the similarity of two spectral bands $k \in (1, N)$ and $l \in (1, N)$, the bigger the value of $C^{k, l}$, which is upper bounded by $1.0$.
It is worth noting that our NBS effectively learns the weight $\alpha$ of each spectral band without retraining and, more importantly, can infer any number $M \in (1, N)$ of chosen spectral bands in one-shot by simple post-processing. As described in Alg. \ref{postprocess}, in each iteration, the post-processing chooses the band according to the highest value in $\alpha$ and suppresses the weight of similar bands after each selection (see Fig. \ref{method_fig}).

\begin{algorithm}
\SetAlgoLined
\SetKwData{Left}{left}\SetKwData{This}{this}\SetKwData{Up}{up}
\SetKwFunction{Union}{Union}\SetKwFunction{FindCompress}{FindCompress}
\SetKwInOut{Input}{input}\SetKwInOut{Output}{output}
\Input{Multiple numbers of spectral bands to select $\{M_1, M_2, \dots\}$; The learned band weights $\alpha$; The band-wise correlation matrices $C \in R^{N \times N}$; The hyper-parameter $\beta$;
}
 \BlankLine
 
 \For{$M$ in $\{M_1, M_2, \dots\}$}{
  \For{$j$ in $(1, M)$}{
    Select the spectral band $k$ corresponding to the largest alpha value in $\alpha$ \;
    
    Compute the new $alpha$ by $\alpha = \alpha \times (1.0 - C^{k})^{\beta}$ \;
  }
 }
 \caption{The post processing procedures of NBS.}
 \label{postprocess}
\end{algorithm}

\textbf{Spectral-wise NBS.}
Current bands search methods exploit the spatial-spectral features for spectral recovery. To investigate the impact of the spatial-wise features, we also present a spectral-wise NBS baseline by completely removing the spatial information. We adopt a seq2seq model that adopts two GRUs for encoding and decoding as the search model to recovery full bands $S(u,v)$ at each pixel $(u,v)$ following the same padding strategy as NBS. The post-process algorithm is the same as Alg. \ref{postprocess}.

\textbf{$M$-equal split NBS.}
Moreover, to further demonstrate the effectiveness of the global band search design of NBS, we provide the $M$-equal split NBS that selects $M$ spectral bands by first splitting all spectral bands $B_i$ to $M$ equal splits. 
Then, the priority of each band within each split is learned following Eq. \ref{darts_optim}. All spectral bands within the sample split are summed after the softmax operation. 
After that, the sums of each split are concatenated together and are fed into the spectral recovery model. As each split only chooses the most prominent band after training, the $M$-equal split NBS does not need post-processing. It exploits the prior of the spectral recovery task and needs re-training for different $M$s.
The comparison result is shown in Table \ref{main_result}.

\begin{table*}[!ht]
\centering
\resizebox{\linewidth}{!}{
\begin{tabular}{lcccccccccccc}
\hline
Selection method         & \multicolumn{3}{c}{Manual}        & \multicolumn{3}{c}{M-equal split} & \multicolumn{3}{c}{Spectral based} & \multicolumn{3}{c}{NBS (ours)} \\
Spectral wavelength (nm) & \multicolumn{3}{c}{630, 530, 470} & \multicolumn{3}{c}{650, 510, 410} & \multicolumn{3}{c}{700, 590, 400}  & \multicolumn{3}{c}{680, 540, 480}          \\ \hline
SR model                & MRAE       & RMSE      & PSNR     & MRAE       & RMSE      & PSNR     & MRAE       & RMSE       & PSNR     & MRAE          & RMSE         & PSNR        \\
HINet \cite{hu2022hdnet}                   & 0.1147     & 0.0178    & 38.84    & 0.0834     & 0.0167    & 39.20    & 0.1043     & 0.0215     & 37.18    & \textbf{0.0765}        & \textbf{0.0153}       & \textbf{40.44}       \\
MIRNet \cite{zamir2020learning}                  & 0.0574     & 0.0116    & 42.92    & 0.0596     & 0.0131    & 41.45    & 0.0921     & 0.0187     & 38.44    & \textbf{0.0534}      & \textbf{0.0104}        & \textbf{43.39}       \\
MST-L \cite{cai2022mask}                   & 0.0601     & 0.0120    & 42.34    & 0.0583     & 0.0121    & 41.83    & 0.0956     & 0.0193     & 38.16    & \textbf{0.0581}         & \textbf{0.0106}        & \textbf{43.46}       \\
MST++ \cite{cai2022mst++}                   & 0.0615     & 0.0123    & 42.55    & 0.0602     & 0.0127    & 41.59    & 0.0958     & 0.0194     & 38.22    & \textbf{0.0555}         & \textbf{0.0102}      & \textbf{43.42}       \\ \hline
\end{tabular}
}
\caption{Quantitative results on the NTIRE validation set. We compare different band search algorithms across four different spectral recovery networks. Our NBS consistently outperforms current baselines.}
\label{main_result}
\end{table*}

\begin{figure}[]
\centering
\includegraphics[width=0.4\linewidth]{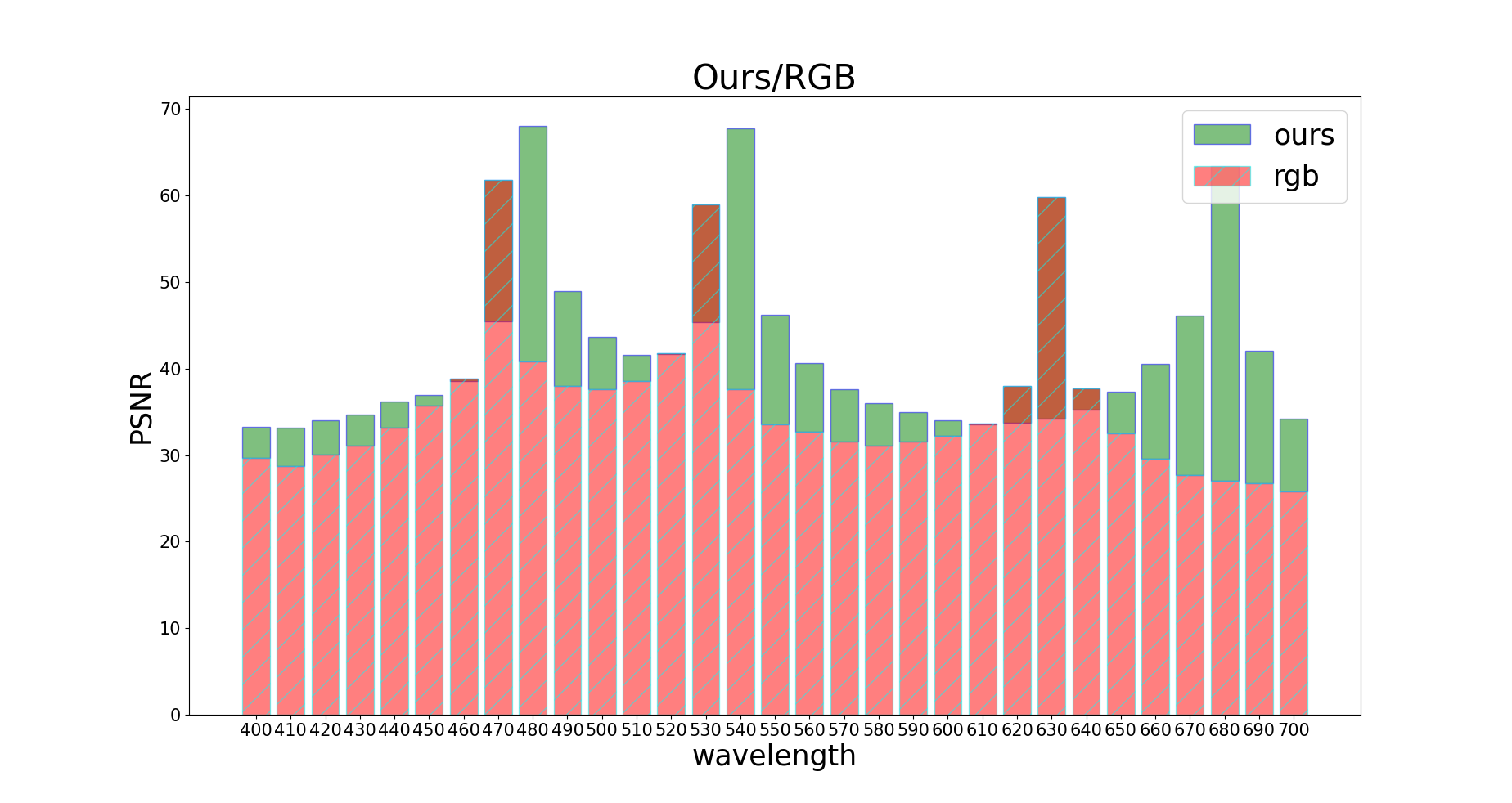}
\caption{The PSNR comparison on the validation set between our NBS searched bands and RGB.
}
\label{error_compare}
\end{figure}

\begin{figure*}[]
\centering
\includegraphics[width=0.6\linewidth]{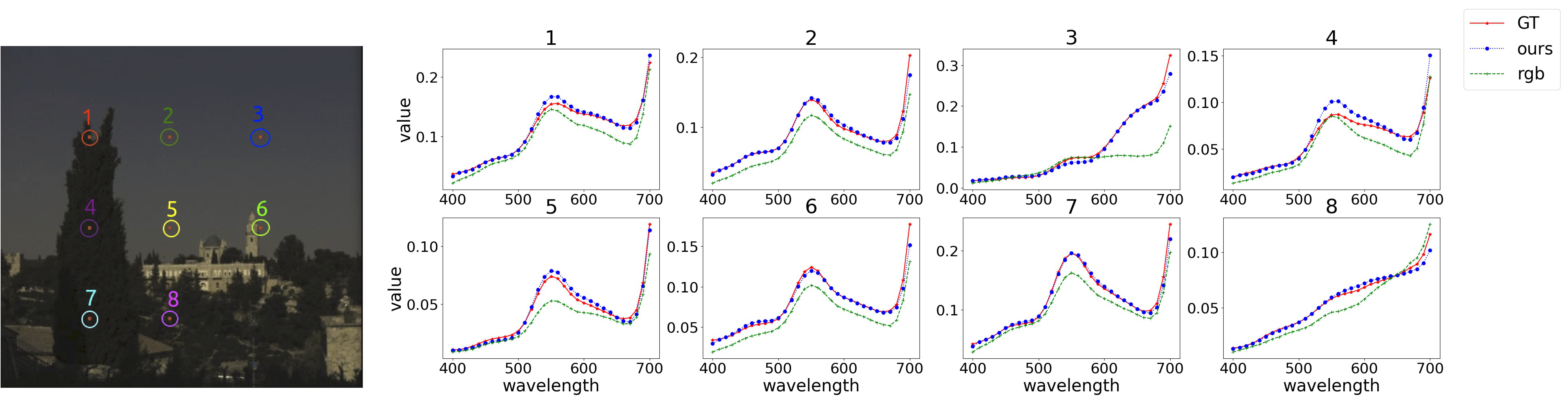}
\caption{Qualitative results on the NTIRE validation set. The spectral recovery result of our searched bands fits the GT better.
}
\label{curve_compare}
\end{figure*}

\section{Evaluations}
In this section, firstly, we show the main results on the NTIRE 2022 spectral reconstruction challenge \cite{arad2022ntire}. Secondly, we present the ablation study of our NBS on the impact of $M$ and $beta$. Lastly, the qualitative results are provided. 

\textbf{Implementation Details.} The NTIRE dataset contains 1000 hyperspectral images. Each image at size of $482 \times 512$ has $31$ wavelengths from $400$ nm to $700$ nm. Current spectral methods like MST++ \cite{cai2022mst++} on the NTIRE dataset focus on recovering the hyperspectral image from three manually selected spectral bands: R ($630$ nm), G ($530$ nm), and B ($470$ nm). The three bands are roughly evenly distributed in all spectral bands. For fair comparisons on different band selection choices, different from MST++, we regard wavelengths at $630, 530, 470$ nm as the RGB inputs rather than the RGB input \cite{cai2022mst++} generated with shot noise.
Four different spectral recovery models (MST++ \cite{cai2022mst++}, MST-L \cite{cai2022mask}, MIRNet \cite{zamir2020learning}, HINet \cite{hu2022hdnet}) are adopted for comparisons in the main results. 

For the search process of NBS, the number of epoch is set to $50$. We also apply the $L2$ regularization with a weight value of $0.01$ to $\alpha$ to avoid overfitting. The hyper-parameter $\beta$ is set to $0.5$ by default and the ablation study in shown in Table \ref{ablation_study}.
For the training of spectral recovery based on selected bands, we set the batch size to $12$ and the default learning rate is $0.0004$. 
The training epochs is $50$ by default and the Cosine Annealing scheme is adopted. Other default training options are the same as MST++ \cite{cai2022mst++}.
Following the NTIRE challenge, we adopt three metrics MRAE, root mean square error (RMSE), and the peak signal-to-noise ratio (PSNR) for comparisons.

\textbf{Main Results.} The main comparisons between different band selection methods on four different spectral recovery models are shown in Table \ref{main_result}. We compare our NBS to the manually selected RGB as well as two strong neural band search baselines (see subsection \ref{nbs_subsection}). The results show that both our special-spectral NBS and the $M$-equal split NBS achieve better reconstruction results than manually selected RGB, showing that the relaxation of the band selection makes our NBS accurately learn the priority of spectral bands. Besides, when using the same HINet model, our default spatial-spectral search method achieves the best PSNR $40.44$, which is $4.1\%$ higher than the RGB counterparts. The PSNR of the spectral-based baseline is at least 8\% lower than our spatial-spectral search result across four SR models, showing that the joint spatial-spectral search is essential for spectral recovery. The PSNR of the M-equal split baseline averaged by four models is $1.66$ lower than our NBS, demonstrating the superior of our global BS framework.
In addition, the per-wavelength PSNR comparison on the whole validation set is plotted in Fig. \ref{error_compare}. Our NBS searched bands show observable gains over RGB on almost all wavelengths. 

\textbf{Impact of $M$ and $beta$.} We compare different combinations of $M$ and $beta$ in Table \ref{ablation_study} to examine their impacts on spectral recovery. With the increase of the number of selected bands, the spectral recovery performance increase significantly, especially from $2$ to $4$. For the choose of $beta$, we examine three values of different magnitudes and empirically find that $0.5$ is a reasonable value which achieves better spectral recovery results.

\begin{table}[!ht]
\centering
\resizebox{0.7\linewidth}{!}
{
\begin{tabular}{clccc}
\hline
\multicolumn{5}{c}{NBS (ours) + MST++}                                                         \\
$M$ & \multicolumn{1}{c}{$\beta$} & MRAE            & RMSE            & PSNR           \\ \hline
3               & 0.01                     & 0.0628          & 0.0128          & 41.79          \\
3               & 0.5                      & \textbf{0.0555} & \textbf{0.0102} & \textbf{43.42} \\
3               & 2.0                      & 0.0685          & 0.0142          & 41.54          \\ \hline
2               & 0.5                      & 0.1233          & 0.0223          & 36.53          \\
4               & 0.5                      & 0.0479          & 0.0081          & 45.26          \\
6               & 0.5                      & 0.0192          & 0.0031          & 53.24          \\
8               & 0.5                      & 0.0111          & 0.0018         & 58.01         \\ \hline \hline
31 (oracle)               & 0.5                      & 0.0019          & 0.0002          & 75.14          \\ 
\hline
\end{tabular}
}
\caption{The impact of $M$ and $\beta$.}
\label{ablation_study}
\end{table}

\textbf{Qualitative Results.} To illustrate the spectral recovery result, we show the qualitative results in Fig. \ref{curve_compare} by randomly selecting a hyperspectral image from the validation set and randomly choosing pixels with equally spaced widths. The full band curve recovered by our NBS fits the ground-truth curve better than RGB with smaller gaps. As our NBS adopts the same spectral recovery model as the RGB counterpart, the small differences demonstrate the effectiveness of our NBS.

\section{Conclusion}
In this paper, we introduced a novel one-shot neural band selection framework for spectral recovery. Different from traditional band search strategies. our NBS is based on the continuous relaxation of the band selection process and allows efficient band search by gradient descent. Our NBS learns the priority of each spectral band in one-shot and can select any number of bands by simple post-processing steps. We also investigated the impact of the spatial features and the effectiveness of our global band search strategy, extensive evaluations also demonstrated the effectiveness of our method.

\vfill\pagebreak
\bibliographystyle{IEEEbib}
\bibliography{refs}

\end{document}